\documentclass[10pt,twocolumn,letterpaper]{article}

\usepackage{wacv}
\usepackage{times}
\usepackage{epsfig}
\usepackage{graphicx}
\usepackage{amsmath}
\usepackage{amssymb}
\usepackage{booktabs}
\usepackage{subfig}
\usepackage[accsupp]{axessibility}

\def\wacvPaperID{2023} 

\wacvapplicationstrack 

\wacvfinalcopy 

\ifwacvfinal
\usepackage[breaklinks=true,bookmarks=false]{hyperref}
\else
\usepackage[pagebackref=true,breaklinks=true,colorlinks,bookmarks=false]{hyperref}
\fi

\begin{document}

\title{Efficient few-shot learning for pixel-precise handwritten document layout analysis}


\author{Axel De Nardin$^1$
,
Silvia Zottin$^1$
,
Matteo Paier$^1$
,
Gian Luca Foresti$^1$
,
Emanuela Colombi$^1$
,\\
Claudio Piciarelli$^1$\\
\and
$^1$University of Udine\\
{\tt\small \{denardin.axel, zottin.silvia, paier.matteo\}@spes.uniud.it}\\
{\tt\small \{gianluca.foresti, emanuela.colombi, claudio.piciarelli\}@uniud.it}
}

\maketitle
\thispagestyle{empty}

\begin{abstract}
    \textit{
   Layout analysis is a task of uttermost importance in ancient handwritten document analysis and represents a fundamental step toward the simplification of subsequent tasks such as optical character recognition and automatic transcription. However, many of the approaches adopted to solve this problem rely on a fully supervised learning paradigm. While these systems achieve very good performance on this task, the drawback is that pixel-precise text labeling of the entire training set is a very time-consuming process, which makes this type of information rarely available in a real-world scenario. In the present paper, we address this problem by proposing an efficient few-shot learning framework that achieves performances comparable to current state-of-the-art fully supervised methods on the publicly available DIVA-HisDB dataset.}
\end{abstract}

\section{Introduction}

Document image layout analysis is a very important task for the humanities community for the study of ancient manuscripts~\cite{simistira2017icdar2017}. In particular, the page segmentation of a given document image into semantically meaningful regions (\eg main text, comments, decorations and background) allows them to easier and quicker study the document and represents a fundamental step toward the simplification of subsequent tasks such as optical character recognition~\cite{ni2017writer} and automatic transcription~\cite{fischer2009automatic}.

Document layout analysis is a particularly challenging task when referring to historical manuscripts. Compared to machine-printed documents~\cite{ramel2007user}, ancient texts exhibit many variations such as layout structure, decorations and different writing styles. For example, in many manuscripts, the main text body is entwined with additions, corrections and marginal or interlinear glosses~\cite{simistira2017icdar2017}, often made by different authors at different times.
Furthermore, historical document pages frequently suffer from high degradation due to aging, ink stains, noise, scratches and bad conservation~\cite{davoudi2021ancient}.
In addition to all these factors, even the image acquisition of ancient text may not be appropriate with illumination issues or inconsistencies and scan curve problems~\cite{baird2003digital}.

Due to the non-uniformity and integrity of the images, many of the approaches adopted to solve this problem rely on a fully supervised learning paradigm~\cite{mehri2015learning, xu2018multi, oliveira2018dhsegment}.
While these systems achieve very good performance on this task, they usually need a large number of annotated images for training. The Ground Truth (GT) represented by these annotations is critical for training and evaluating document analysis methods, especially for complex historical manuscripts that exhibit challenging layouts with interfering and overlapping handwriting~\cite{garz2016creating}.
\begin{figure*}[htb]%
    \centering
    \subfloat[\centering CSG18 page]{{\includegraphics[width=.25\linewidth]{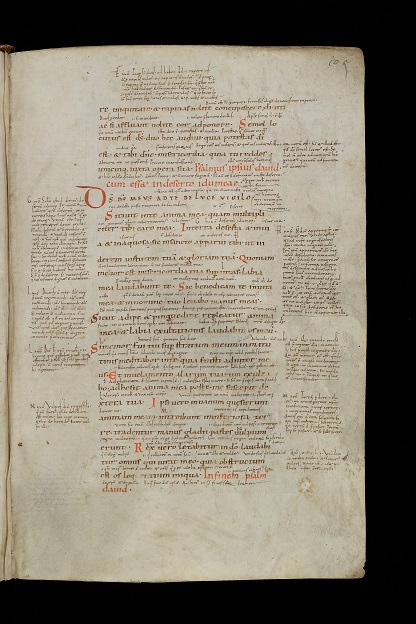} \label{fig:cs18page} }}%
    \qquad
    \subfloat[\centering CSG863 page]{{\includegraphics[width=.25\linewidth]{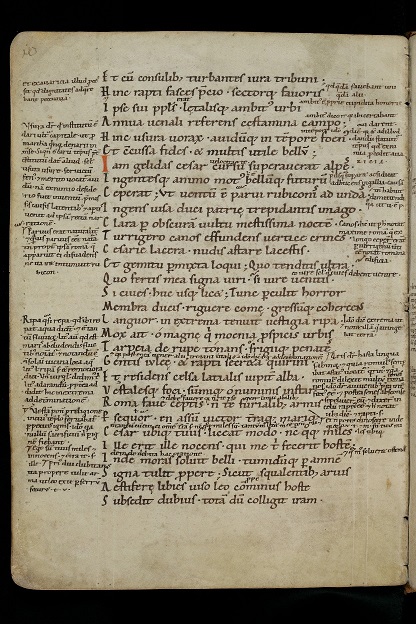} \label{fig:cs863page}}}
    \qquad
    \subfloat[\centering CB55 page]{{\includegraphics[width=.25\linewidth]{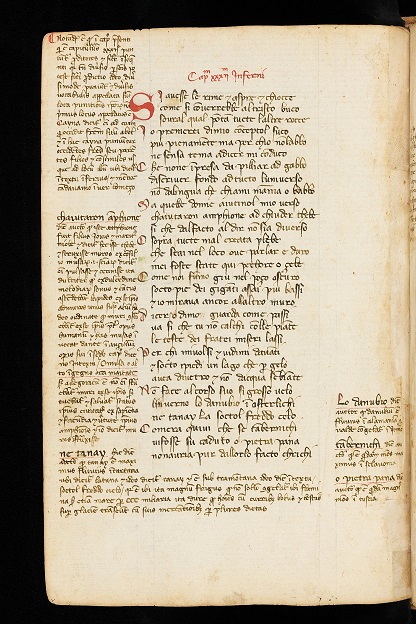} \label{fig:cb55page}}}%
    \qquad
    \subfloat[\centering CSG18 detail]{{\includegraphics[width=.25\linewidth]{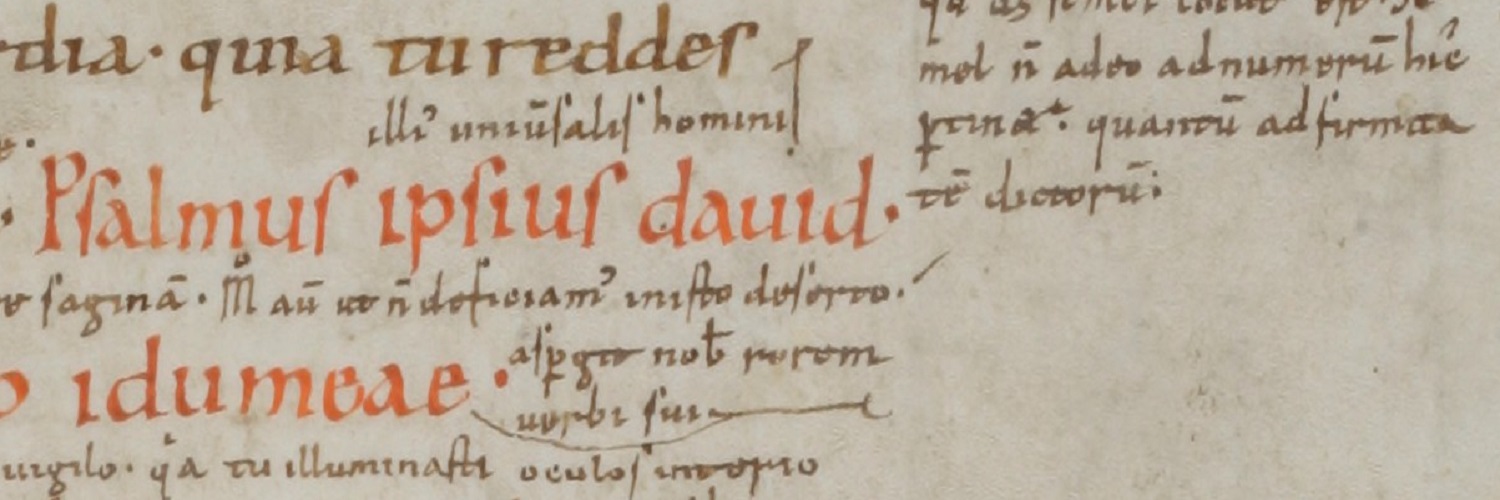} \label{fig:cs18det}}}%
    \qquad
    \subfloat[\centering CSG863 detail]{{\includegraphics[width=.25\linewidth]{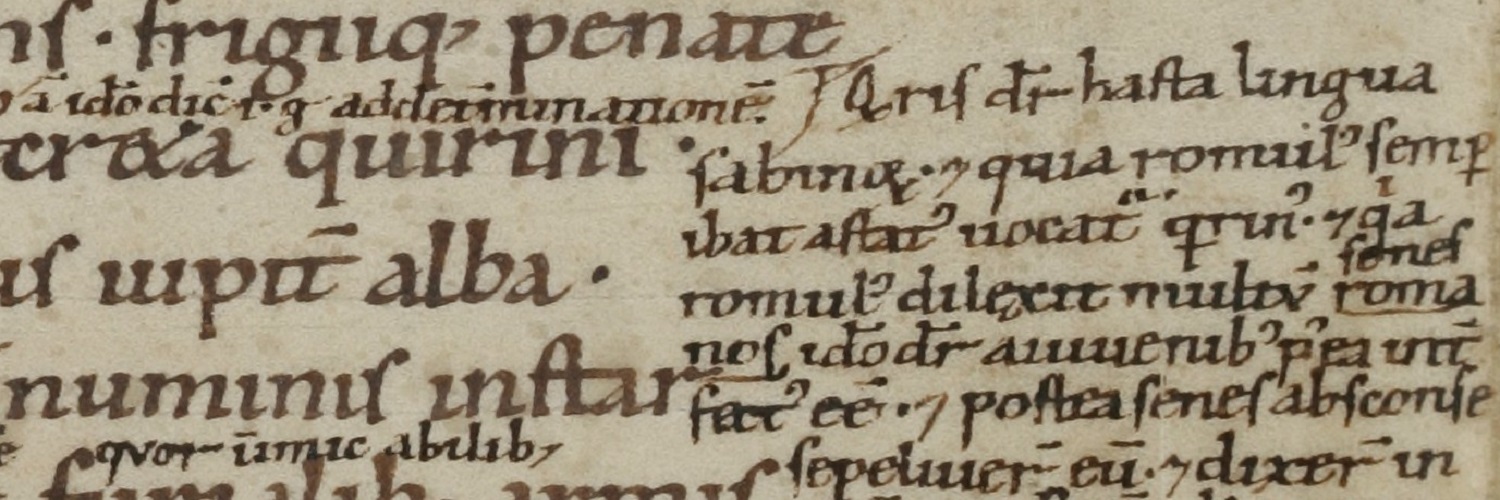} \label{fig:cs863det}}}
    \qquad
    \subfloat[\centering CB55 detail]{{\includegraphics[width=.25\linewidth]{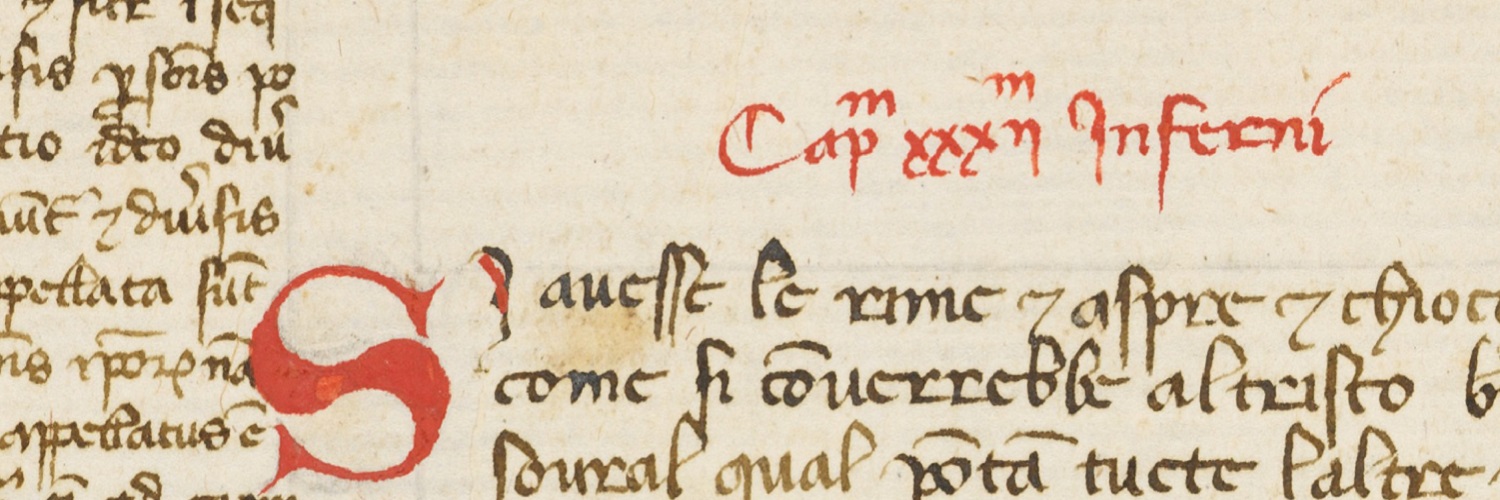} \label{fig:cb55det}}}%
    \caption{Samples from the 3 manuscripts (CSG18, CSG863 and CB55) presents in DIVA-HisDB dataset~\cite{simistira2016diva}. Fig.~\ref{fig:cs18page}\textendash~\ref{fig:cb55page} show a full page for each manuscripts, while Fig.~\ref{fig:cs18det}\textendash~\ref{fig:cb55det} show a detail extracted from each of them.}%
    \label{fig:example}%
\end{figure*}
The drawback is that pixel-precise annotation of the entire dataset of historical document pages requires specific domain knowledge as well as being a very time-consuming process, making this type of information rarely available in a real-world scenario. Nonetheless, few-shot learning approaches are still under-explored in the literature for this task.
This paper tackles all the above issues by proposing a novel few-shot learning framework for efficient pixel-precise layout segmentation of historical documents.
In particular, we propose two original contributions: first, a dynamic instance generation process that aims at providing a way of efficiently leveraging the limited data available in this scenario and second a segmentation map refinement process that provides a way of improving the precision of the annotation predictions provided by the adopted model.
By combining these two components with a powerful DeepCNN backbone network we are able to achieve performances comparable to the ones obtained by current state-of-the-art fully supervised approaches.

The rest of this paper is organized as follows. Section~\ref{relatedwork} gives an overview of some related work in page segmentation for historical document images. Then Section~\ref{proposedmethod} describes the three components defining the proposed framework. Section~\ref{experiments} reports the details of our experimental setup as well as providing an overview of the obtained results. Finally, in Section~\ref{conclusions}, are drawn the conclusions of this work and discuss the ideas for future work.

\section{Related work} \label{relatedwork}

Many different approaches have been proposed to tackle the layout analysis, especially for handwritten historical documents. This section reviews some representative state-of-the-art methods for historical document image segmentation.
In general, the techniques employed for document layout analysis are usually divided into three categories: bottom-up, top-down and hybrid~\cite{binmakhashen2019document}.

The bottom-up strategy derives document analysis dynamically from smaller granularity data levels such as pixels and connected components. Then, the analysis grows up to form larger document regions and stops once it reaches a page segmentation into different regions with uniform elements.
These techniques are flexible and do not require any prior knowledge of the layout structure. However, usually, they demand many labeled training data that is often not available, especially in the domain of historical documents where highly specialized expertise is needed to label the data.

On the contrary, top-down approaches assume that pages have a well-defined structure and layout. Various characteristics of the document page structure are then considered, such as white space between text regions, size of text blocks and the measures between main texts and paratext~\cite{davoudi2021ancient}.
The page segmentation process then starts from the whole page and cuts it into areas to produce small homogeneous regions.
In general, the top-down methods are easily applicable but not suitable for complex layouts such as handwritten historical documents. In addition, these methods depend on the layout structure of the document, so they have a low generalization capability.

Even though the research of this technique is well established, there are still many challenging issues that neither bottom-up nor top-down strategies can address appropriately. For this reason, the hybrid strategy has been identified and derives from the integration of the other two main categories~\cite{binmakhashen2019document}.
Over the years, many techniques have been used to address this task, from classical computer vision algorithms to deep learning methods.

Chen \etal~\cite{chen2015page} used a convolutional autoencoder to learn the features directly from the pixel intensity values. Then, by using these features to train a Support Vector Machines (SVM), this method got high-quality segmentation without any assumption of specific topologies and shapes of document layouts.

A different approach, which also allows performing layout analysis, was proposed by Mehri \etal~\cite{mehri2015learning} with the method based on learning texture features. This method used the simple linear iterative clustering super-pixels, Gabor descriptors, the co-occurrence matrix of the gray level, and a SVM to classify pixels into foreground and background. A super-pixel is a set of pixels that shares similar spatial and intensity information.

Many researchers have approached the page segmentation problem as a pixel labeling problem such as the work by Chen \etal~\cite{chen2016page}. In this paper, the features are learned directly from randomly selected image patches by using stacked convolutional autoencoders. With a SVM trained with the features of the central pixels of the super-pixels, an image is segmented into four regions. Finally, the segmentation results are refined by a connected components-based smoothing procedure. The authors show that by using super-pixels as units of labeling, the speed of the method is increased.

Following the same idea of \cite{chen2016page}, in Chen \etal~\cite{chen2016random} local features are learned with stacked convolutional autoencoders in an unsupervised manner for the purpose of initial labeling. Then a conditional random field model is applied for modeling the local and contextual information jointly to improve the segmentation results. The graph nodes are represented by super-pixels, so the label of each pixel is determined by the label of the super-pixels to which it belongs.
\begin{figure}[htb]%
    \centering
    \subfloat[\centering Original page]{{\includegraphics[width=.48\linewidth]{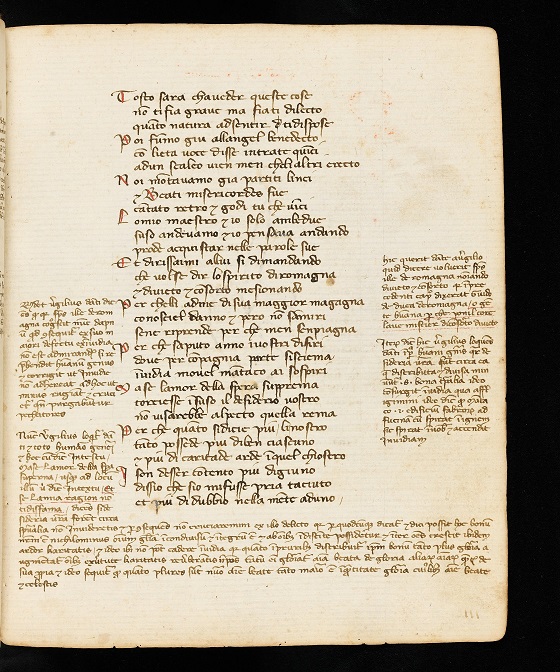} \label{fig:sauvolaimg} }}%
    \subfloat[\centering Masked page]{{\includegraphics[width=.48\linewidth]{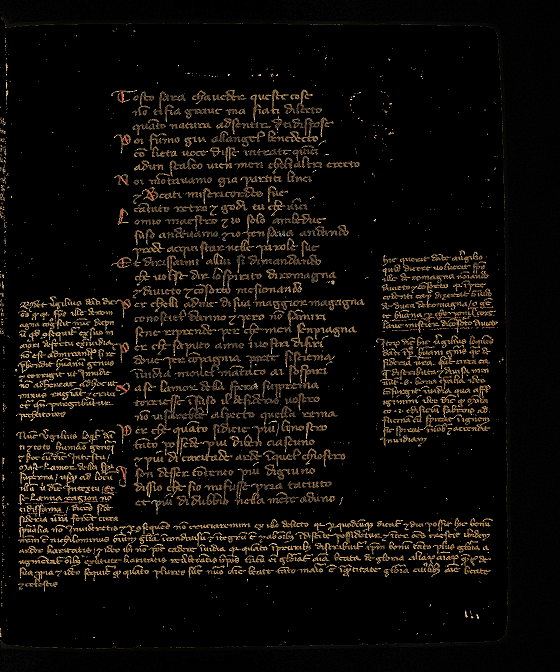} \label{fig:sauvolamask}}}
    \caption{Image showing a sample instance from the CB55 manuscript class \ref{fig:sauvolaimg} together with the corresponding filtered version \ref{fig:sauvolamask} obtained by masking it with the corresponding binarized filter extracted from it through the Sauvola thresholding algorithm.}%
    \label{fig:sauvola}%
\end{figure}
Another interesting approach was proposed by Xu \etal~\cite{xu2018multi} with a multi-task layout analysis framework based on the fully convolutional network to solve the page segmentation, text line segmentation and baseline detection problem simultaneously. The framework trains a multi-task fully convolutional network to predict pixel-wise classes and heuristic-based post-processing is adopted to reduce noise and correct misclassification. The prediction of the four branches was combined to produce the result of page and text line segmentation.

Davoudi \etal~\cite{davoudi2021ancient} proposed a novel method for document layout analysis that reduces the need for labeled data. This method is a dictionary-based feature learning model where a sparse autoencoder is first trained in an unsupervised manner on a historical text document’s image patch. The latent representation of image patches is then used to classify pixels into various region categories of the document using a feed-forward neural network.

Finally, Studer \etal~\cite{studer2019comprehensive} tackles the problem of the limited presence of annotated data by introducing and testing the use of pre-train models on images from a different domain and then fine-tuning them on historical documents. The authors choose some famous, pre-trained, semantic-segmentation networks on the ImageNet~\cite{imagenet} database for object recognition (\eg DeepLabV3~\cite{deeplab} and SegNet~\cite{segnet}) and test them on the task of text segmentation in handwritten documents. The results demonstrated that on some manuscripts pre-training on ImageNet increases the performance, but on others, the pre-trained network performs much worse.
\section{Proposed Method} \label{proposedmethod}
In this section, we provide an outline of the proposed framework showcasing the key components that define its effectiveness. First, we introduce DeepLabV3, a robust CNN architecture for segmentation tasks which we employed as the backbone of our framework. Then we discuss the dynamic instance generation approach that characterizes our training process, which allows for improved generalization capabilities at a low computational cost. Finally, we introduce the refinement process we applied to the segmentation maps produced by the backbone network to obtain a more precise version of them.

\begin{figure*}[htb]
    \centering
    \includegraphics[width=.97\linewidth]{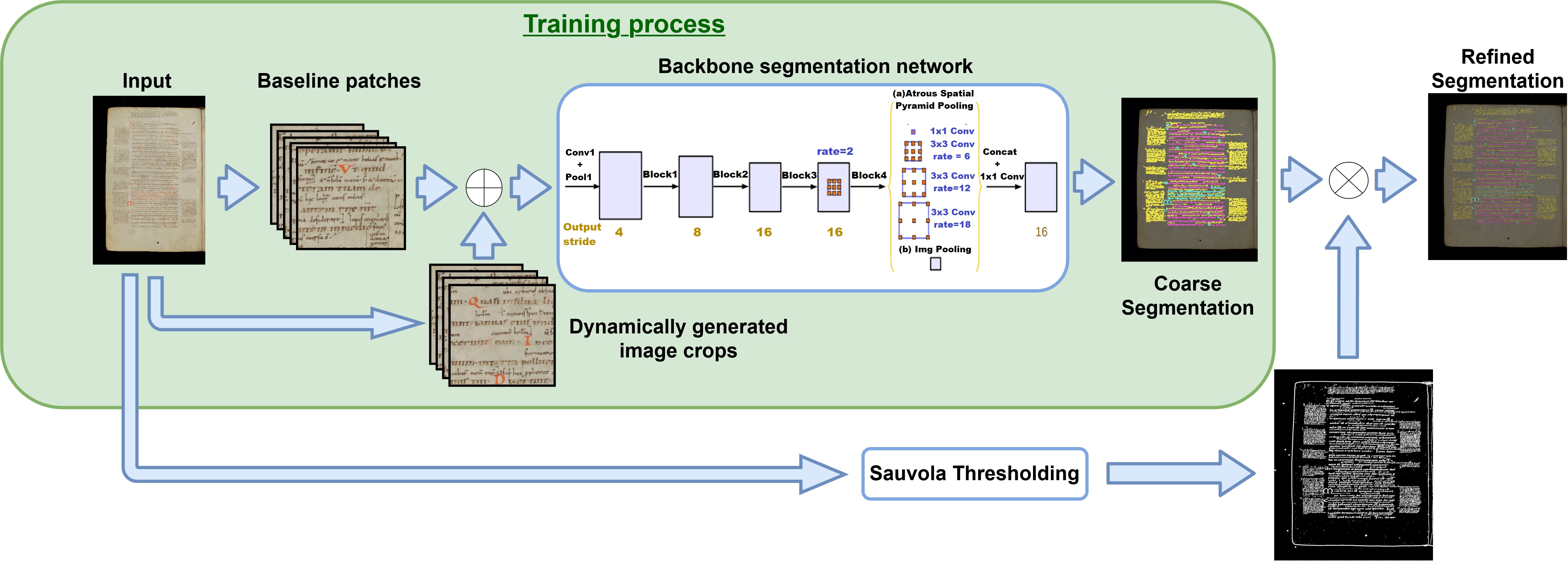}
    \caption{Visual representation of the segmentation pipeline for the proposed framework. During the training (green area) each input image is split into N, non-overlapping, patches of size $k\times k$ which cover its entire surface and are used as the baseline training set. Furthermore, during each epoch $C$ random crops of the input image are randomly generated from the image as additional training data. The backbone model then provides a predicted coarse segmentation map for each of these patches that are compared with the ground truth one through the application of a weighted cross-entropy loss. At inference time the dynamic instance generation step is removed while a segmentation refinement process is applied to the outputs of the backbone architecture to obtain more precise segmentation maps.}
    \label{fig:segpipeline}
\end{figure*}

A visual representation of the presented framework is provided in Fig.~\ref{fig:segpipeline}. 

\subsection{Backbone network}
As the backbone of the proposed framework, we selected DeepLabV3~\cite{deeplab} , a ResNet-based Deep CNN architecture that employs atrous (dilated) convolutions in cascade or in parallel with different dilation levels and is widely adopted in the context of image semantic segmentation. This approach allows retaining a larger spatial resolution for the feature maps throughout the network architecture compared to models relying heavily on striding and pooling layers.

The key aspect that makes atrous convolutions effective for in the context of semantic segmentation task is that they allow us to create deeper networks while at the same time providing output feature maps that are larger than those of a traditional deep CNN architecture and without any increase in the amount of computation needed.
Furthermore, the Atrous Spatial Pyramid Pooling (ASPP) module employed in the DeepLabV3 network allows, thanks to the adoption of different dilation rates, provides an effective way of capturing information at multiple scales of the original image.

\subsection{Dynamically enhanced training data}
\begin{figure}[htb]%
    \centering
    \subfloat[\centering Baseline patches]{{\includegraphics[width=.39\linewidth]{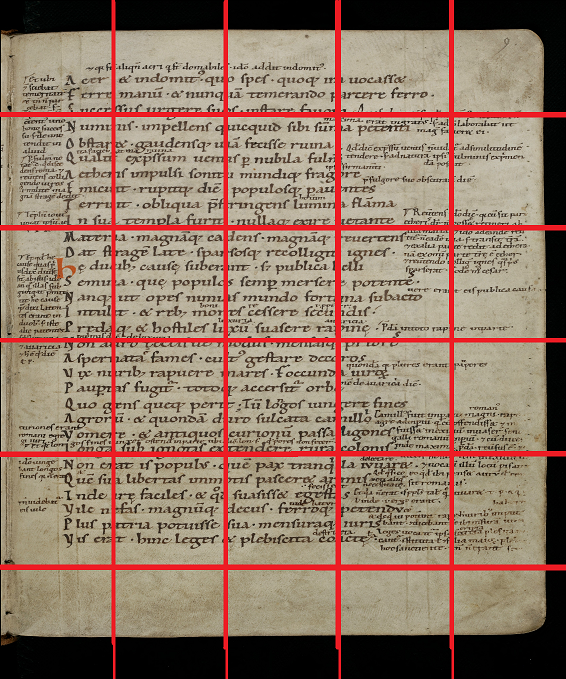} \label{fig:baselinepatches} }}%
    \subfloat[\centering Randomly selected crops]{{\includegraphics[width=.39\linewidth]{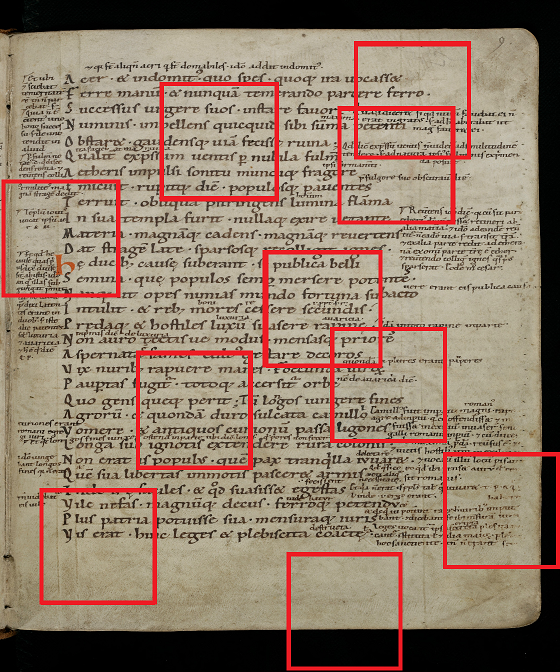} \label{fig:randomcrops}}}
    \caption{Representaton of the instance generation process: \ref{fig:baselinepatches} shows the baseline non overlapping patches, while \ref{fig:randomcrops} shows a set of randomly generated crops.}%
    \label{fig:trainsetenhancement}%
\end{figure}

Maximizing the exploitation of available data is a key step in few-shot learning systems like the one we are presenting in this work. While using the entire images to train our model would allow capturing global contextual information about it, we believe that most of this contextual information can also be retrieved from smaller sections of the document's pages almost as effectively. For this reason, as a first step to improve the efficiency of our training setup, we decided to split each page of the document in a set $P$ of non-overlapping, fixed-size, patches that cover the entire input image and represent our baseline training set (Fig.~\ref{fig:baselinepatches}).
While this first step allows us to enhance the size of our training set by a factor of $P$ its main limitation is that the size of the single patches must be large enough to allow capturing contextual information from the corresponding represented area of the original image, thus limiting the value of $P$. 
For this reason, to further improve the efficiency of our training process while at the same trying to enhance the generalization capabilities of our model we introduced a dynamic instance generation process which at each epoch retrieves a set of $C$ randomly selected crops, of the same size as the baseline dataset patches, which, together with the corresponding segmentation maps, are used as additional instances for training the segmentation network (Fig.~\ref{fig:randomcrops}). An additional perk of this approach is that it avoids generating huge amounts of potentially unnecessary data beforehand. In particular, the intuition behind this strategy is that we try to generate a number of instances that is proportional to the complexity of the dataset. A complex dataset will need a larger amount of instances to be generated to fully represent the heterogeneity of its data, while for a simpler dataset a smaller number of them would suffice.  Furthermore, being employed only during training allows for maintaining a lean pipeline at inference time when only the baseline patches are employed for the segmentation process.

\subsection{Segmentation refinement}
One common problem for semantic segmentation systems in the context of handwritten document analysis is represented by the degree of precision that the systems need to achieve in order to provide pixel-precise segmentations of the foreground elements of the document's pages (\eg text, decorations). What makes this task even more difficult in historical documents is that they are subjected to varying degrees of degradation which both reduce the contrast between the foreground and the background and also can lead to having less clearly shaped characters.

To solve this problem we introduce in our inference pipeline a segmentation refinement process based on the Sauvola thresholding technique~\cite{SAUVOLA2000225}. Sauvola's  thresholding can be seen as an improvement over Niblack's algorithm~\cite{niblack} which was specifically designed for document binarization. This technique computes a local threshold $t$ at each pixel of a grayscale image by applying the Eq.~\ref{eq:sauvola} to the $n \times n$ surrounding pixels defining its neighborhood:
\begin{equation}
    t =mN * (1+k*((stdN/R)-1))
    \label{eq:sauvola}
\end{equation}
In the above equation, $mN$ and $stdN$ represent the mean and the standard deviation of the neighborhood respectively, while $R$ represents the dynamic range of standard deviation. Finally, $k$ is a constant value that controls the value of the threshold in the local window, the higher is k the lower the threshold from the local mean.
In practice, this process provides a binary mask that allows to effectively separate the darker areas of the image, representing the background, from the lighter ones, representing the foregrounds. The segmentation refinement process is carried out by performing a pixel-wise multiplication between the segmentation maps provided by the backbone architecture (where the background class is represented as 0) and the mask extracted from the corresponding input image by the Sauvola algorithm. In Fig.~\ref{fig:sauvola} we show a sample page from the dataset together with its filtered version.

\section{Experiments} \label{experiments}
In this section, we provide an overview of the dataset by highlighting its characteristics as well as the challenges it poses, together with a detailed description of the training setup adopted for the experiments.

\subsection{Dataset}
\begin{figure}[htb]%
    \centering
    \subfloat[\centering Original page]{{\includegraphics[width=.48\linewidth]{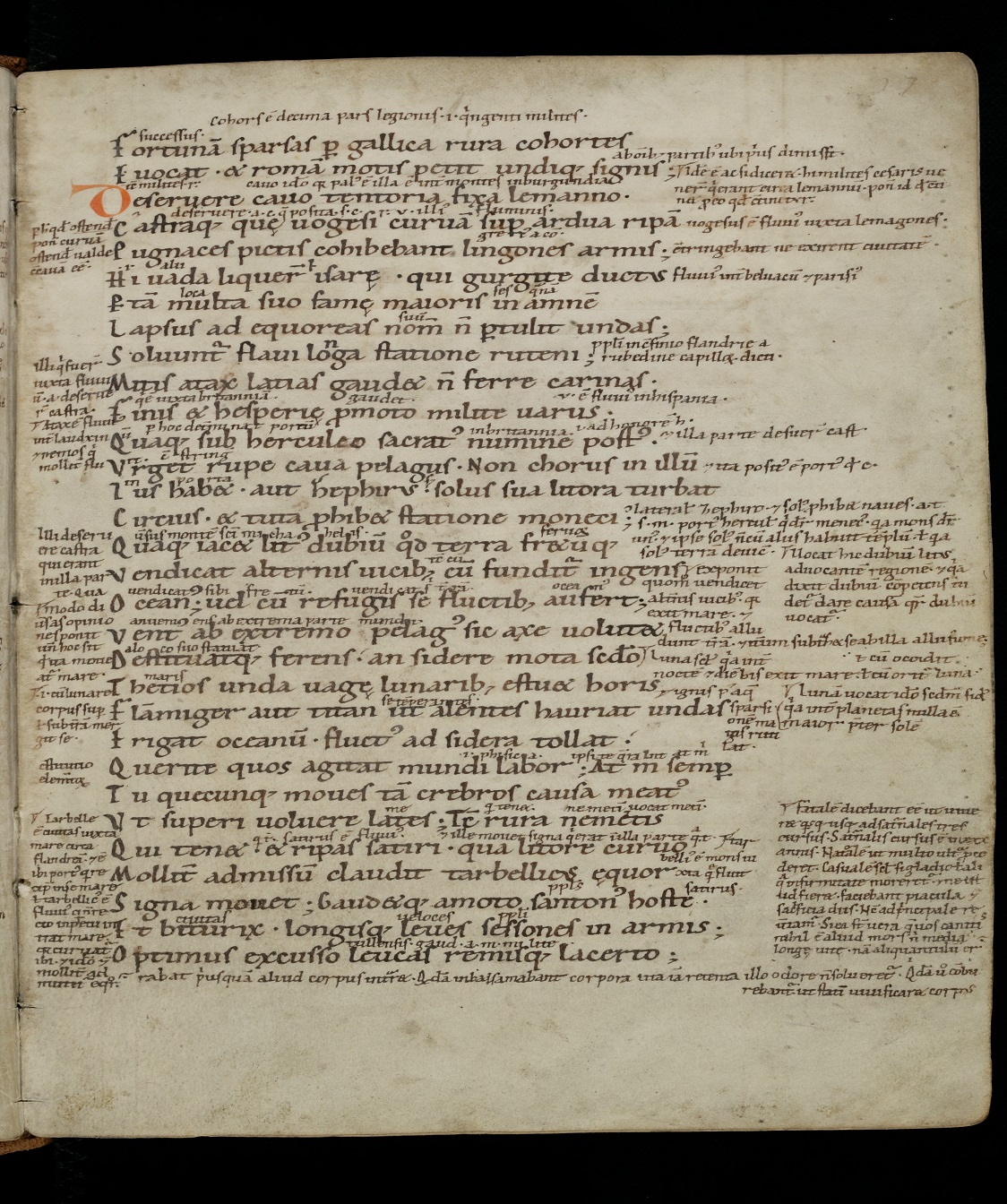} \label{fig:gtimage} }}%
    \subfloat[\centering Page GT]{{\includegraphics[width=.48\linewidth]{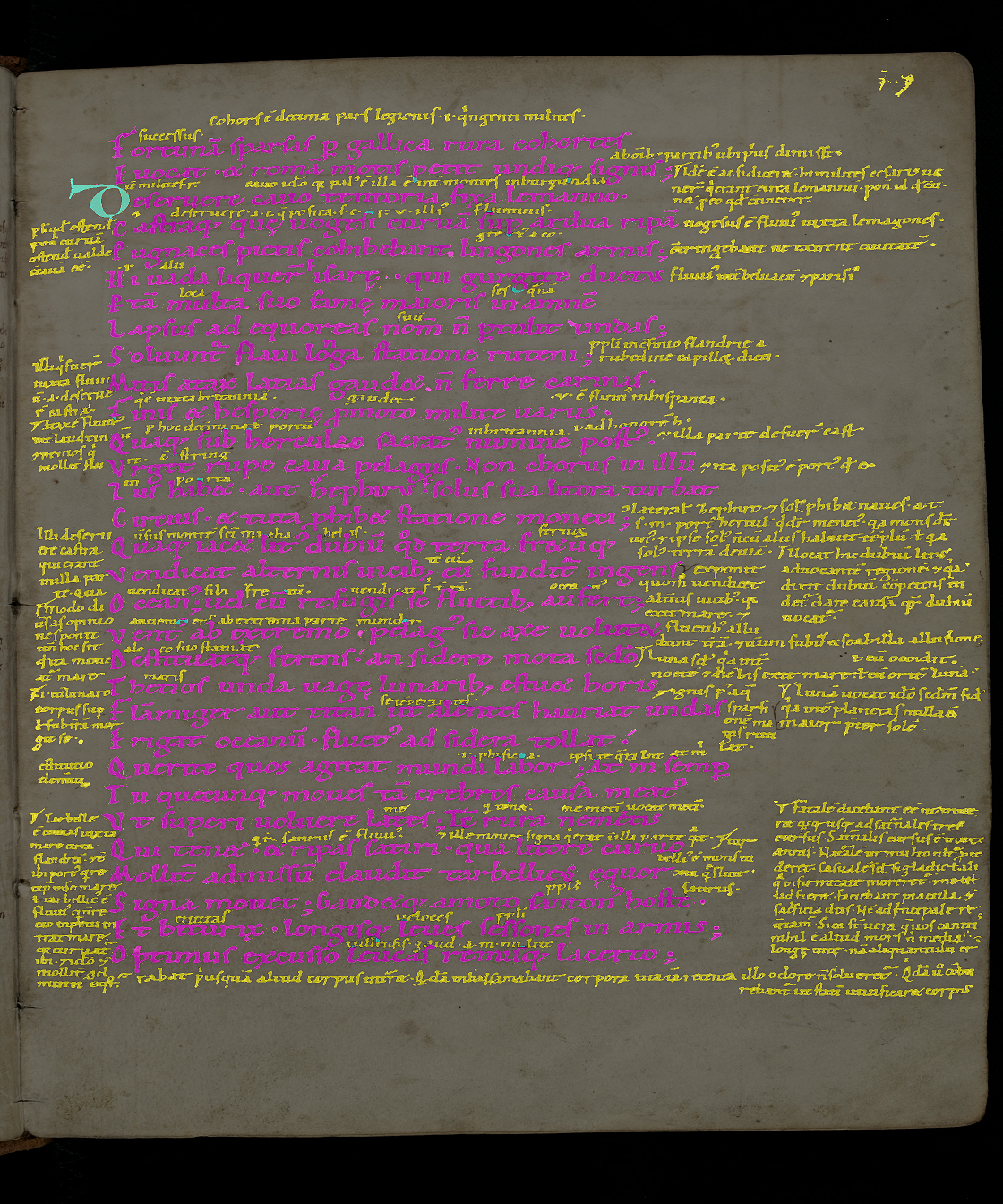} \label{fig:gtmask}}}
    \caption{Images showing: A page of the CSG863 manuscript class (~\ref{fig:gtimage}) and the corresponding ground truth mask (~\ref{fig:gtmask}) 
    The magenta areas represent the main text, while the yellow and cyan area represent the comments and decorations respectively.}%
    \label{fig:groundtruth}%
\end{figure}

The dataset selected to train and test our system is the DIVA-HisDB dataset~\cite{simistira2016diva}, a historical document dataset consisting of a total of 150, high-resolution, annotated pages coming from 3 different medieval manuscripts, identified as CSG18, CSG863 and CB55, characterized by complex and heterogeneous layouts as well as different levels of degradation. A sample page for each of the classes of manuscripts is reported in Fig.~\ref{fig:example}. Out of the total images 60 are typically used for training, 30 for validation and another 60 for testing. For the present work, we only relied on 6 images (2 for each manuscript) to train our model.
The DIVA-HisDB provides pixel-level ground truth segmentation (Fig.~\ref{fig:groundtruth}) for the layout of each page, which distinguishes between 4 classes of elements. A further challenging aspect of the dataset is a very marked imbalance between these classes. Detail of their distribution is provided in Tab.~\ref{tab:classdistr}.

\begin{table}[htb]
\begin{center}
    {\small{
\begin{tabular}{ccccc}
\hline
                & \textbf{BG} & \textbf{Comments} & \textbf{Decoration} & \textbf{Text} \\ \hline
\textbf{CB55}   & 82.41               & 8.36              & 0.55                & 8.68               \\
\textbf{CSG18}  & 85.16               & 6.78              & 1.47                & 6.59               \\
\textbf{CSG863} & 77.82               & 6.35              & 1.83                & 14.00              \\ \hline
\end{tabular}}}
\end{center}

\caption{Classes distribution (\%).}
\label{tab:classdistr}
\end{table}

\subsection{Training and inference setup}
The network training process has been carried out through the adoption of the ADAM optimizer with a learning rate of $1e^-3$ and a weight decay of $1e^-5$. The selected loss function is a weighted cross-entropy loss in which the weights for each class are determined by taking the square root of 1 over the class frequency in the dataset (Eq.~\ref{eq:weights}, where $F_i$ represents the frequency (\%) of class in the corresponding class dataset), this choice was made to account for the class imbalance that characterizes the dataset.
\begin{equation}
    W_i = \sqrt{\frac{1}{F_i}}
    \label{eq:weights}
\end{equation}
The maximum number of epochs was set to 200, starting from epoch 50 onwards an early stop was introduced in case the network did not improve over the last 20 iterations.
Due to the high resolution of the images (up to $4.8k \times 6.8k$ px), a resizing process has been carried out to reduce the computational complexity of the model and be able to fit them in the GPU memory. The final shape of the images is of $1120\times 1344$ px. To train the model two images for each manuscript class have been selected and split into patches of size $224\times 224$ px, resulting in a training set of 60 patches for each manuscript. This set is then enhanced by generating 10 additional random crops of the same size for each image as part of our dynamic training routine. The maximum final amount of generated instances composing the training set is 4000 if the model needs all the available epochs to converge.
As for the inference setup the parameters selected for the Sauvola thresholding algorithm where a window size of 15pixels and $k = 0.1$, the reason behind the selection of such a low value for the latter parameter is that we wanted to avoid as much as possible including background noise, represented by page degradation, as part of the generated binary mask.

\subsection{Experimental Results}
In this section, we outline the metrics selected for the evaluation of the proposed framework as well as provide both an ablation study, aimed at supporting the effectiveness of the choices defining the proposed system and a thorough comparison with other popular semantic segmentation approaches, including the current state-of-the-art for document layout analysis.

\subsection{Metrics}
The metrics used to evaluate the performance of the proposed approach are Precision, Recall, Intersection over Union (IoU) and F1-Score. The metrics were calculated as described in \cite{xu2018multi} individually for each class following the definition reported in Eq.~\ref{prec}\textendash~\ref{f1}, where TP, FP and FN stand respectively for True Positives, False positives and False Negatives, and then a weighted average, based on each class frequency has been performed.
\begin{align}
    &Precision = \frac{TP}{TP+FP} \label{prec}\\
    &Recall = \frac{TP}{TP+FN} \label{rec}\\
    &IoU = \frac{TP}{TP+FP+FN} \label{iou}\\
    &F1-score = \frac{2 \times Precision \times Recall}{Precision + Recall} \label{f1}
\end{align}
    
\subsection{Results}
\begin{figure}[htb]%
    \centering
    \subfloat[\centering Ground truth]{{\includegraphics[width=.95\linewidth]{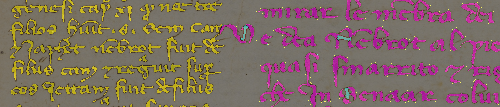} \label{fig:refprocgt} }}%

    \subfloat[\centering Coarse Prediction]{{\includegraphics[width=.95\linewidth]{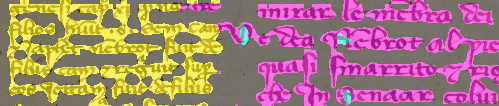} \label{fig:refproccoarse}}}

    \subfloat[\centering Refined Prediction]{
    {\includegraphics[width=.95\linewidth]{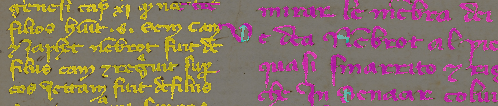} \label{fig:refprocrefined}}}\\

    \caption{Qualitative results showing the effects of the segmentation refinement process. Fig.~\ref{fig:refprocgt} shows the original ground truth for a zoomed area of the original image. Fig.~\ref{fig:refproccoarse} shows the coarse segmentation mask obtained by the model. Finally Fig.~\ref{fig:refprocrefined} shows the segmentation prediction resulting from the refinement process.}%
    \label{fig:refinementprocess}%
\end{figure}

First we present the results of the ablation study we conducted on the different versions of the proposed framework. In particular, in Tab.~\ref{tab:ablation} we provide a comparison between the baseline approach, which consists in running the backbone network on a patch level but without the dynamic crop generation nor the segmentation refinement processes, and the improved versions in which these strategies have been employed singularly and the in a combined fashion. We report both the scores for the individual classes as weel as the final averaged ones. As we can see both the proposed strategies determine an improvement in the system performance when introduced as part of the segmentation pipeline. In particular, the segmentation map refinement process, for which we show a sample qualitative results in Fig.~\ref{fig:refinementprocess}, provides the most important contribution, improving the mean scores for all the metrics by a margin of 5.5\% to 9\%. The dynamic crop generation other hand provides an additional average boost of 2\% on the selected metrics. Finally, when combined, the 2 strategies provide an average improvement of 9\% across the metrics, with a peak for the IoU score which gets improved by a 12.3\%. The values in bold represent the best-performing system for each metric on the corresponding class of the dataset.

\begin{table*}[htb]
\Huge
\centering
\resizebox{.97\textwidth}{!}{
\begin{tabular}{lllllllllllllllll}
\hline
 &
  \multicolumn{4}{c}{\textbf{CB55}} &
  \multicolumn{4}{c}{\textbf{CSG18}} &
  \multicolumn{4}{c}{\textbf{CSG863}} &
  \multicolumn{4}{c}{\textbf{Mean}} \\ \hline
&
  Prec & Rec & IoU & F1 & Prec & Rec & IoU & F1 & Prec & Rec & IoU & F1 & Prec & Rec & IoU & F1 \\ \hline
\textit{Ours (baseline)} &
  0.846 & 0.843 & 0.757 & 0.825 & 0.911 & 0.918 & 0.851 & 0.904 & \textbf{1.000} & 0.913 & 0.913 & 0.954 & 0.919 & 0.891 & 0.840 & 0.894 \\
\textit{Ours (w/ seg. refinement)} &
  0.962 & 0.930 & 0.913 & 0.945 & 0.982 & 0.978 & 0.964 & 0.980 & \textbf{1.000} & 0.913 & 0.913 & 0.954 & \textbf{0.981} & 0.940 & 0.930 & 0.960 \\
\textit{Ours (w/ dynamic crop gen.)} &
  0.902 & 0.908 & 0.833 & 0.896 & 0.935 & 0.939 & 0.893 & 0.933 & 0.914 & 0.915 & 0.844 & 0.906 & 0.917 & 0.921 & 0.857 & 0.912 \\
\textit{Ours (w/ both)} &
  \textbf{0.974} & \textbf{0.974} & \textbf{0.950} & \textbf{0.972} & \textbf{0.985} & \textbf{0.982} & \textbf{0.968} & \textbf{0.982} & 0.985 & \textbf{0.984} & \textbf{0.971} & \textbf{0.984} & \textbf{0.981} & \textbf{0.980} & \textbf{0.963} & \textbf{0.980}\\
  \hline
\end{tabular}}
\caption{Results of the ablation study. Each rows shows the performance of the different versions of our system across all the selected metrics for the 4 classes of manuscripts composing the DIVA-HisDB dataset. The last four columns show the average scores achieved by the models across the different classes.}
\label{tab:ablation}
\end{table*}

Hereafter, we compare the results of the proposed approach with 5 popular semantic segmentation models which proved to be effective on a wide variety of segmentation tasks across different domains, namely the original version of DeepLabV3~\cite{deeplab}, its improvement, represented by DeepLabV3+ ~\cite{deeplabv3plus}, FCN~\cite{fcn}, Lite Reduced Atrous Spatial Pyramid Pooling (LRASPP)~\cite{lraspp} and Pyramid Scene Parsing Network (PSPNet)~\cite{pspnet}. Furthermore, we compare our model with the current state-of-the-art model for layout analysis of ancient documents, which we will refer to as MLA~\cite{xu2018multi}.
All the models, excluding MLA for which we gathered the results from the respective paper, have been personally tested by us keeping the training and evaluation settings as consistent as possible.
\begin{figure*}[htb]
    \centering
    \includegraphics[width=.98\linewidth]{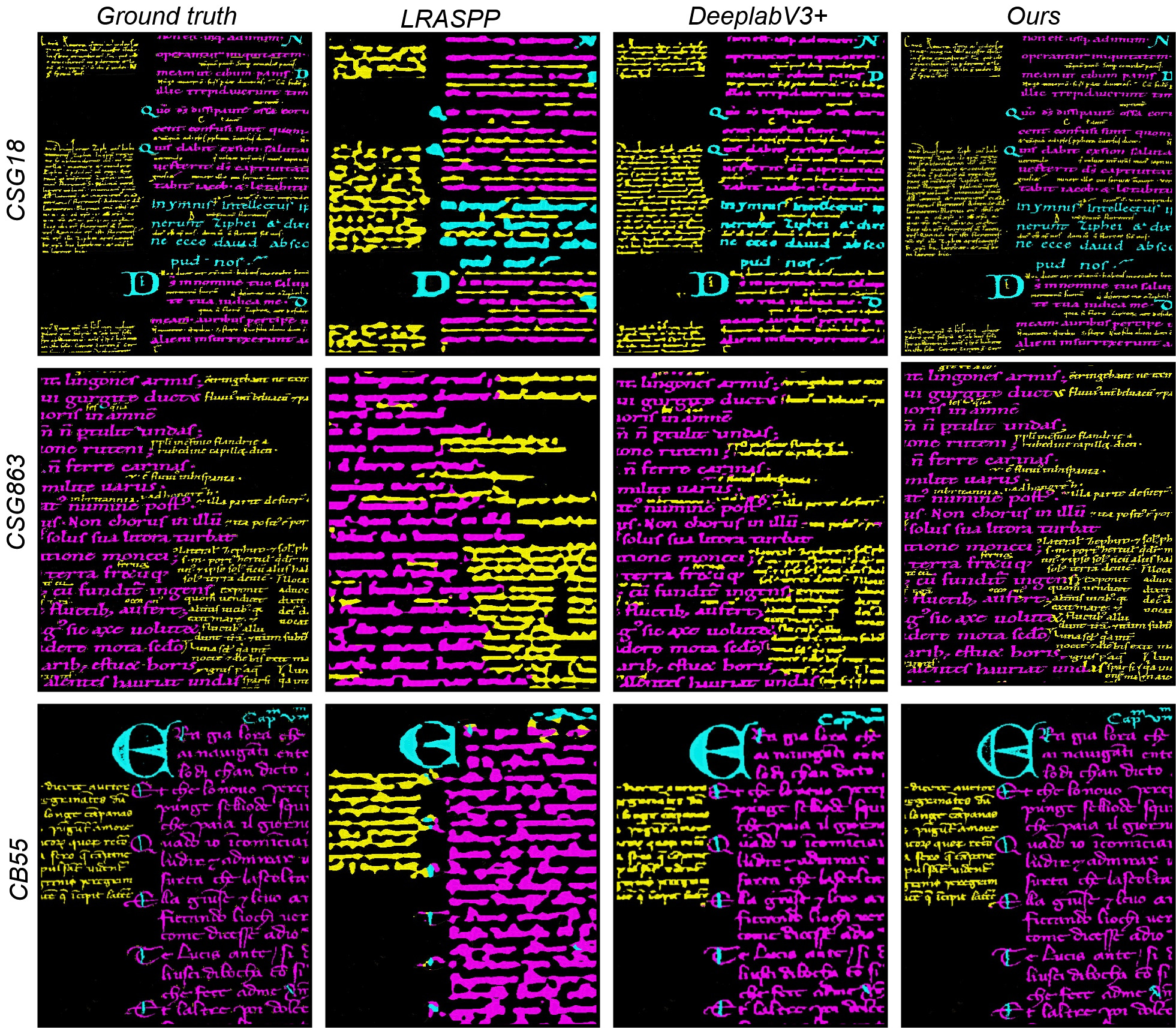}
    \caption{Image showing a qualitative comparison between our framework and the competition ones. Each row represents a zoomed area belonging to a different instance of the dataset, representing the three classes of manuscript contained in it. In the first column, the ground truth segmentation maps for the 3 images are shown, while on the remaining columns we provide the results produced by the three systems, LRSAPP, DeepLabV3+ and Ours respectively.}
    \label{fig:qualitativeres}
\end{figure*}
In Tab.~\ref{tab:results} we provide the final results obtained by the aforementioned models. As we can see all the reported approaches achieve scores above 90\% for the Precision and Recall, and F1-Score and above 84\% for the IoU metric, which is arguably the most interesting metric to consider while being also the hardest on which to consistently achieve good results. Nonetheless, we show how the framework proposed in the present paper is able to consistently and substantially improve the results obtained by the other approaches, with the exclusion of MLA, while relying on a fraction of the available training data. In particular, the most impactful improvement has been achieved for the IoU metric on which our system was able to outperform the second best performing model from the competition, DeepLabV3+, by a 4.3\% margin, with an average improvement across all the metrics of 2.9\%.

Furthermore, when compared to MLA, which being the state-of-the-art represents the most interesting term of comparison, our approach was able to effectively close the gap to it, reaching as close as a 0.8\% gap for the Precision metric, while being outperformed by only a 1.6\% margin on average across all the metrics. It is important to keep in mind though that MLA is a much heavier system, which was trained over 180.000 patches extracted from all the images of the available training set, while the proposed approach relied on at most 4000 patches extracted from just 2 of the available images for its training process, resulting in a reduction of the data needed by a factor of 45.

Finally, in Fig.~\ref{fig:qualitativeres} we provide a set of qualitative results for the proposed framework as for the DeepLabV3+ and LRASPP models. In particular, we show a comparison between the segmentation maps produced by the three models and those representing the ground truth on a zoomed section of three different images belonging to the three classes of manuscripts present in the dataset. 
As we can see the three models provide different levels of precisions when it comes to the fidelity to the ground truth segmentation maps. The maps produced by LRASPP, while identifying correctly the different foreground components of the page, are still very coarse and include a wide area of background around as well as inside the different characters. Deeplabv3+ provides much more precise masks, especially for the main text (magenta areas) for which it is able to correctly identify the boundaries of the characters, while it still struggles with the comments and decorations areas. Finally, we show that the approach presented in this paper is able to correctly identify all the three classes of foreground components while achieving a high degree of precision for all of them across the different types of manuscripts.

\begin{table}[htb]
\huge
\centering
\resizebox{.45\textwidth}{!}{
\begin{tabular}{lllll}
\hline
 & Precision & Recall & IoU & F1 \\ \hline
FCN \cite{fcn}  & 0.918 & 0.916 & 0.843 & 0.904 \\
LRSAPP \cite{lraspp} & 0.930 & 0.911 & 0.854 & 0.910 \\
PSPNet \cite{pspnet} & 0.904 & 0.910 & 0.838 & 0.899 \\
Deeplabv3 \cite{deeplab} & 0.918 & 0.915 & 0.842 & 0.903 \\
Deeplabv3+ \cite{deeplabv3plus} & 0.958 & 0.956 & 0.920 & 0.954 \\
MLA \cite{xu2018multi} & \textbf{0.989} & \textbf{0.995} & \textbf{0.989} & \textbf{0.995} \\ \hline
\textbf{Ours} & \underline{0.981} & \underline{0.980} & \underline{0.963} & \underline{0.980} \\ \hline
\end{tabular}}
\caption{Comparison between the results obtained by our model and the competition, the best and second best score for each metric are the bold and the underlined ones respectively.}
\label{tab:results}
\end{table}

\section{Conclusions} \label{conclusions}
In this paper, we presented a complete framework for few-shot, pixel-precise semantic segmentations on handwritten historical documents. In particular, we introduced two new components which represent an effective way of addressing relevant challenges in the context of few-shot learning systems. These components are a dynamic instance generation module which provides an effective solution to the problem of efficient training sample leveraging when only a small set of training data is available to the system, and a segmentation refinement module through which we were able to consistently retrieve substantially more precise segmentation maps compared to the baseline output of the backbone segmentation network.
By combining these components into the final segmentation framework we showed how it was able to achieve substantially peter performance than other popular semantic segmentation approaches trained on the entire dataset in a traditional setting, as well as being able to achieve results comparable to the current state-of-the-art for historical document text segmentation for the selected metrics on the challenging DIVA-HisDB dataset.

For future works, we would like to address the drawback of the current approach represented by the need to manually select additional parameters for the segmentation refinement process by exploring automated solutions that we believe solve this problem while at the same time further improving the quality of the framework's predictions.
{\small
\bibliographystyle{ieee_fullname}
\bibliography{main}
}

\end{document}